\documentclass[10pt,conference]{IEEEtran}
\IEEEoverridecommandlockouts

\usepackage{cite}
\usepackage{amsmath,amssymb,amsfonts}
\usepackage{algpseudocode} 
\usepackage{graphicx}
\usepackage{textcomp}
\usepackage{xcolor}
\usepackage{algorithm} 
\def\BibTeX{{\rm B\kern-.05em{\sc i\kern-.025em b}\kern-.08em
    T\kern-.1667em\lower.7ex\hbox{E}\kern-.125emX}}

\usepackage{pbalance}

\usepackage{fancyhdr}

\fancypagestyle{arxivcopyright}{
    \fancyhf{} 
    
    \fancyfoot[C]{\footnotesize 
    © 2026 IEEE. Personal use of this material is permitted. Permission from IEEE must be obtained for all other uses, in any current or future media, including reprinting/republishing this material for advertising or promotional purposes, creating new collective works, for resale or redistribution to servers or lists, or reuse of any copyrighted component of this work in other works.}
}

\makeatletter
\newcommand{\linebreakand}{%
  \end{@IEEEauthorhalign}
  \hfill\mbox{}\par
  \mbox{}\hfill\begin{@IEEEauthorhalign}
}
\makeatother

\begin{document}
\title{An Inclusive and Lightweight Approach to Federated Continual Learning for Cultural Heritage
}

\author{
\IEEEauthorblockN{
Ioannis Theologitis$^{1}$,
Debin Meng$^{2}$,
Stylianos Eleftheriadis$^{1}$,
Vasileios Lolis$^{1}$,
Konstantinos Votis$^{1}$
}
\IEEEauthorblockA{
\small $^1$Information Technologies Institute, Centre for Research and Technology Hellas, Thessaloniki, Greece\\
\small $^2$School of Electronic Engineering and Computer Science, Queen Mary University of London, UK\\
{\tt\small \{theolo, stylelev, vaslwlis, kvotis\}@iti.gr}\\
{\tt\small debin.meng@qmul.ac.uk}
}
}

\maketitle


\vspace{-1em} 
\begin{center}
    \footnotesize\itshape
    This is the accepted author manuscript of a paper accepted at the
    2026 IEEE International Conference on Cyber Humanities (IEEE-CH 2026), Venice, Italy, September 7--9, 2026. \\
    © 2026 IEEE. Personal use of this material is permitted. Permission from IEEE must be obtained for all other uses, in any current or future media, including reprinting/republishing this material for advertising or promotional purposes, creating new collective works, for resale or redistribution to servers or lists, or reuse of any copyrighted component of this work in other works.
\end{center}
\vspace{1em}

\begin{abstract}
Artificial intelligence can support cultural heritage and digital humanities through large-scale retrieval and analysis of digitized collections. However, cultural heritage data are often distributed across institutions, constrained by ownership and access restrictions, and continuously evolving over time. Federated Continual Learning (FCL) is well suited to this setting, as it enables models to learn from distributed and sequential data without sharing raw collections. In this paper, we propose FedCurv-DR, a lightweight, regularisation-based FCL strategy. The method accumulates parameter-importance estimates across clients and experiences to protect learned knowledge, while updating them only at fixed intervals to minimize communication and computation overhead. We evaluate FedCurv-DR in a continual learning scenario using the WikiArt image dataset for genre classification with evolving styles, reporting performance, energy, and fairness metrics. Our results show that FedCurv-DR reduces forgetting and balances performance, fairness, and energy efficiency for sustainable AI in cultural heritage.
\end{abstract}


\section{Introduction}

AI is increasingly adopted across many domains, including cultural heritage~\cite{fiorucci_machine_2020}. However, cultural heritage data raise specific ethical and technical challenges. Collections are often distributed across institutions and continuously evolve through digitisation. Digitisation processes may also reproduce existing societal and cultural biases, including the under-representation or misrepresentation of minority groups. These biases can then be inherited or amplified by AI pipelines trained on such collections~\cite{foka_tracing_2025,zhitomirsky-geffet_what_2022}. Such bias can derive from unequal access to cultural heritage data, since ownership, copyright, licensing, and privacy constraints can limit the availability of collections for computational research and AI development~\cite{disli2025copyright}. At the same time, the unequal capacity of institutions to participate in digitisation and AI adoption can further reinforce this bias, as many cultural heritage organisations face limited funding, insufficient technical expertise, and resource-intensive digitisation requirements~\cite{eprs2023aiheritage}. 

\begin{figure*}[t]
\centerline{\includegraphics[width=0.85\textwidth,height=0.28\textheight,keepaspectratio]{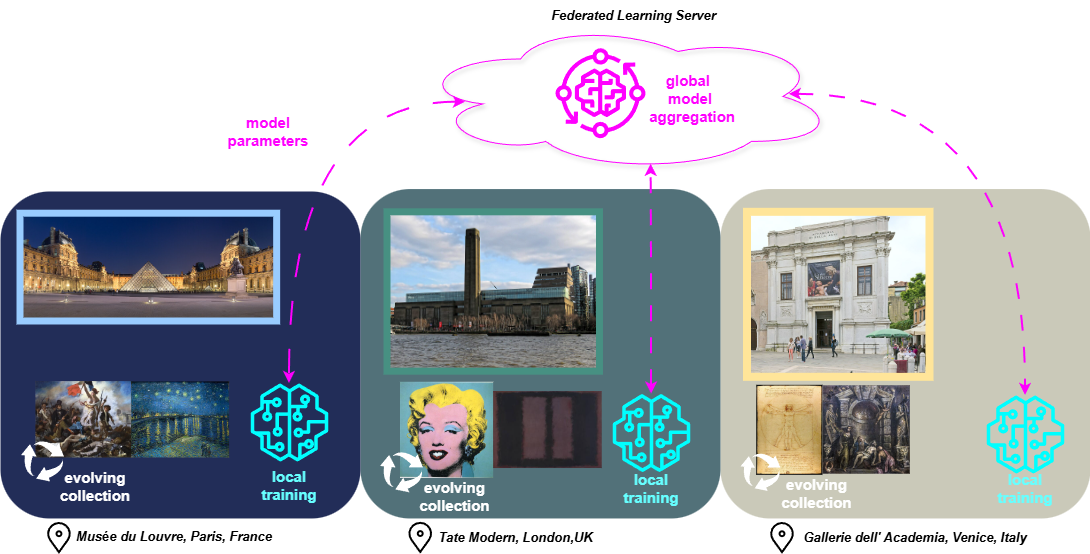}}
\caption{Overview of the federated continual learning process. Art institutions across Europe collaborate on training a single global model for image classification on their evolving collections, without sharing private data. Institutions periodically send model updates, rather than raw data, to a server operated by a trusted coordinating party, which aggregates them into a shared global model.}
\label{fig:fcl_overview}
\end{figure*}

These challenges highlight the need for AI solutions that are green, trustworthy and inclusive - core principles of the Cyber Humanities vision, which calls for cultural heritage technologies that combine technical innovation with ethical governance and equitable participation~\cite{adorni_towards_2025,bellini_cyber_2025}.
Federated Learning (FL) enables collaborative model training without sharing raw data~\cite{mcmahan_communication-efficient_2023}, while Continual Learning (CL) enables models to adapt to sequentially arriving data without retraining from scratch. Their combination, Federated Continual Learning (FCL), can support green AI objectives and is well suited to cultural heritage settings~\cite{tziolas_energy-efficient_2026}. As shown in Fig.~\ref{fig:fcl_overview}, art institutions across the world can collaboratively train AI models on their private and evolving collections while respecting data sensitivity. However, to ensure equal participation in this process, energy efficiency and fairness must be considered when deploying such systems.

In this work, we propose a lightweight, exemplar-free FCL framework for cultural heritage image classification. We introduce FedCurv-DR, a regularisation-based strategy that periodically aggregates parameter-importance information across clients to adapt to new data, or experiences, while preserving performance on past data without relying on replay buffers or computationally intensive methods such as synthetic data generation. We evaluate FedCurv-DR on WikiArt, a large-scale public art dataset relevant to cultural heritage and digital humanities. By treating genre classification as the fixed objective and introducing artistic styles sequentially as new experiences, we model a realistic domain-incremental scenario where digitised cultural collections evolve over time.

The main contributions of this paper are:
\begin{itemize}
    \item We propose FedCurv-DR, a lightweight regularisation-based FCL strategy suitable for cultural heritage.
    \item We evaluate the method on the public WikiArt image dataset for genre classification with style-based continual experiences.
    \item We report performance, energy, and fairness metrics to evaluate the sustainability of our method in accordance with Green AI principles~\cite{schwartz2020greenai}.
\end{itemize}

\section{Related Works}

\subsection{Federated Learning}

In Federated Learning (FL), multiple clients collaboratively train a shared global model under the coordination of a central server, while keeping their data local and private. Instead of exchanging raw data, clients train models locally and periodically send model updates to the server for aggregation. The server-side aggregation algorithms used to combine these updates are commonly referred to as FL strategies. A representative baseline FL strategy is FedAvg~\cite{mcmahan_communication-efficient_2023}, which constructs the global model by averaging client parameters weighted according to the sizes of the local datasets.

Despite its advantages for privacy preservation and scalability, FL introduces several key challenges. In real-world settings, client data are typically not independent and identically distributed (non-IID). In cultural heritage, for example, the number of artifacts may vary substantially between national museums and local galleries, while semantic distributions may differ across styles, periods, or art movements. This heterogeneity can cause local models to diverge during training, as each client optimizes toward its own data distribution, thereby hindering the convergence of the global model. This phenomenon is commonly referred to as inter-client drift~\cite{schoinas_federated_2024}. Several FL strategies have been proposed to address this challenge under heterogeneous data conditions~\cite{li_federated_2020,lin_ensemble_2021,shoham_overcoming_2019,yao_continual_2020,asad_fedopt_2020}.

However, while many of these methods aim to improve performance under non-IID conditions, they may still favor larger or more representative clients, leading to models that perform better for dominant data distributions while neglecting under-represented participants. This can introduce or reinforce fairness concerns, particularly when performance disparities across clients are not explicitly reported. Several works have therefore investigated fairness in FL and proposed fairness-aware strategies to mitigate client-level bias~\cite{li_fair_2020,ezzeldin_fairfed_2022,zhang_towards_2025,lo_fairwag_2025,zhang_unified_2024}.

\subsection{Continual Learning}
Continual Learning (CL) addresses the problem of learning from a stream of tasks or data distributions over time, without retraining from scratch. A central challenge in this setting is catastrophic forgetting~\cite{kirkpatrick_overcoming_2017}, where the model rapidly loses previously acquired knowledge when trained on new tasks. To mitigate this, several families of methods have been proposed. Regularization-based approaches such as Elastic Weight Consolidation (EWC)~\cite{kirkpatrick_overcoming_2017}, Memory Aware Synapses (MAS)~\cite{aljundi_memory_2018}, and Synaptic Intelligence (SI)~\cite{zenke_continual_2017} estimate the importance of model parameters for past tasks and penalize changes to those parameters during subsequent training. Alternatively, replay-based methods~\cite{lopez-paz_gradient_2022,rolnick_experience_2019} maintain a buffer of past examples, allowing the model to rehearse previous knowledge while learning new information. Replay methods have been shown to outperform regularization-based methods in various continual learning scenarios~\cite{ven_three_2019}, but they raise scalability and privacy-related issues, as past data need to be stored.

\subsection{Federated Continual Learning}

Federated Continual Learning (FCL) combines the challenges of Federated Learning and Continual Learning, requiring models to simultaneously cope with data heterogeneity across clients and the sequential arrival of new data or experiences. In this setting, both inter-client drift and catastrophic forgetting must be mitigated, making the learning process significantly more complex. Several approaches have been proposed to address these challenges~\cite{yoon_federated_2021,usmanova_federated_2022,zhang_target_2023}.

Despite their effectiveness, many FCL methods introduce additional computational and communication overhead, for example through synthetic data generation~\cite{zhang_target_2023}. These factors can significantly increase energy consumption, particularly in large-scale or resource-constrained deployments, and may even make such methods infeasible in practice. This is a critical consideration for an inclusive cultural heritage AI ecosystem. In contrast to distillation- or generation-based FCL methods, our approach does not require replay data, memory buffers, or computationally intensive mechanisms. Instead, it relies on parameter-importance regularization with reduced communication and computational overhead.

\section{Methods}
In order to achieve lightweight and exemplar-free federated continual learning, we enable continual learning during local training. This led us to explore FedCurv~\cite{shoham_overcoming_2019}, combined with recent findings on the EWC strategy~\cite{liu_elastic_2026}. Our method reduces the communication and computation costs of the original FedCurv strategy and repurposes it to mitigate catastrophic forgetting.

\subsection{Elastic Weight Consolidation}

Elastic Weight Consolidation (EWC) was proposed to mitigate catastrophic forgetting~\cite{kirkpatrick_overcoming_2017}. In this method, after training on each experience, a Fisher information matrix is calculated using the current training dataset. This matrix represents the importance of each model parameter for the corresponding experience. The diagonal Fisher importance for parameter $\theta_i$ after experience $t$ is approximated as

\begin{equation}
F_{t,i}
=
\frac{1}{|\mathcal{D}_{t}|}
\sum_{(\mathbf{x},y)\in \mathcal{D}_{t}}
\left(
\frac{\partial}{\partial \theta_i}
\log p_{\boldsymbol{\theta}}(y|\mathbf{x})
\right)^2 .
\label{eq:fisher}
\end{equation}

In subsequent experiences, this matrix is used to compute a penalty term for the training loss. The EWC objective is defined as

\begin{equation}
\mathcal{L}_{\mathrm{EWC}}(\boldsymbol{\theta})
=
\mathcal{L}_{\mathrm{task}}(\boldsymbol{\theta};\mathcal{D}_{t})
+
\frac{\lambda}{2}
\sum_i
F_{t-1,i}
\left(
\theta_i - \theta^{*}_{t-1,i}
\right)^2 ,
\label{eq:ewc_loss}
\end{equation}

where $\mathcal{L}_{\mathrm{task}}$ is the standard task loss, $\lambda$ controls the strength of the regularization, and $\boldsymbol{\theta}^{*}_{t-1}$ denotes the model parameters after learning the previous experience. This term protects previously learned knowledge by penalizing changes to important model parameters, encouraging the optimizer to update less important parameters when learning future experiences.

In the original version of EWC, a different Fisher matrix is calculated for each experience, which can introduce scalability issues. Online EWC improves memory efficiency by aggregating Fisher matrices using a decay factor that controls the contribution of past experiences to the current penalty:

\begin{equation}
\bar{F}_{t,i}
=
\beta \bar{F}_{t-1,i}
+
F_{t,i},
\label{eq:online_fisher}
\end{equation}

where $\bar{F}_{t,i}$ is the accumulated importance estimate and $\beta \in [0,1]$ is the decay factor.

\subsection{FedCurv}

FedCurv~\cite{shoham_overcoming_2019} adapts regularization-based continual learning strategies, such as EWC, to the federated learning setting. However, it is designed to address client heterogeneity rather than catastrophic forgetting. FedCurv proposes summing Fisher matrices across clients after each training round into a single global Fisher matrix, which is then used to constrain local updates toward a global consensus and mitigate client drift.

Let $\mathcal{S}_{r}$ be the set of clients selected at communication round $r$, and let $F^{r}_{k,i}$ denote the Fisher importance of parameter $\theta_i$ computed by client $k$. The server aggregates the client importances as

\begin{equation}
F^{r}_{G,i}
=
\sum_{k \in \mathcal{S}_{r}}
F^{r}_{k,i}.
\end{equation}

Clients use this global Fisher matrix during local training to apply the EWC penalty in \eqref{eq:ewc_loss}. In this way, they avoid changing parameters that are important for other clients' distributions, helping the model achieve stable convergence in non-IID settings.

\subsection{Logit Inversion}

In recent work~\cite{liu_elastic_2026}, the authors propose EWC-DR, an improvement to the Fisher matrix calculation in EWC. They observe that, when the model assigns high confidence to the correct class, the standard Fisher estimate in \eqref{eq:fisher} may become small, leading to under-protection of parameters that are nevertheless important. To address this issue, they propose inverting the logits before computing the importance estimates. Let $\mathbf{z}_{\boldsymbol{\theta}}(\mathbf{x})$ denote the logits of the model. The inverted logits are defined as
$\tilde{\mathbf{z}}_{\boldsymbol{\theta}}(\mathbf{x}) = -\mathbf{z}_{\boldsymbol{\theta}}(\mathbf{x})$ and the corresponding predictive distribution is $\tilde{p}_{\boldsymbol{\theta}}(y|\mathbf{x})
=
\mathrm{softmax}
\left(
\tilde{\mathbf{z}}_{\boldsymbol{\theta}}(\mathbf{x})
\right)_y .
\label{eq:inverted_softmax}
$
The Fisher importance is then computed using this modified distribution:

\begin{equation}
\tilde{F}_{t,i}
=
\frac{1}{|\mathcal{D}_{t}|}
\sum_{(\mathbf{x},y)\in \mathcal{D}_{t}}
\left(
\frac{\partial}{\partial \theta_i}
\log \tilde{p}_{\boldsymbol{\theta}}(y|\mathbf{x})
\right)^2 .
\label{eq:dr_fisher}
\end{equation}

As a result, \eqref{eq:dr_fisher} assigns higher importance values to parameters associated with highly confident predictions.

\subsection{Our Method (FedCurv-DR)}

We adapt FedCurv from a client-drift regularization method into a federated continual learning method. We make three main modifications. First, we incorporate the EWC-DR improvement into the calculation of parameter importances. Each client computes $\tilde{F}^{r}_{k,i}$ using \eqref{eq:dr_fisher}. Second, we modify the importance aggregation logic by accumulating importances not only across clients, but also across experiences. We also include a decay factor to reduce redundant protection of outdated experiences. The global importance estimate is updated as

\begin{equation}
\bar{F}^{r}_{G,i}
=
\beta \bar{F}^{r-1}_{G,i}
+
\sum_{k \in \mathcal{S}_{r}}
\tilde{F}^{r}_{k,i},
\label{eq:ours_global_fisher}
\end{equation}
where $\bar{F}^{r}_{G,i}$ is the accumulated global importance of parameter $\theta_i$ at round $r$, and $\beta$ is the decay factor.
Third, we introduce an interval term $I$, following the example of \cite{yao_continual_2020}, that controls how often importances are calculated and transmitted, reducing communication and computation costs.

The pseudocode for our method is shown in Alg.~\ref{alg:fedcurv_dr}. It is worth noting that, for $\beta = 0$ and $I = 1$, our method is equivalent to the aggregation logic of the original FedCurv.

\begin{algorithm}[t]
\caption{FedCurv-DR}
\label{alg:fedcurv_dr}
\begin{algorithmic}[1]
\Statex \textbf{Input:} rounds $R$, clients $K$, interval $I$, regularization strength $\lambda$, decay factor $\beta$, datasets $\{\mathcal{D}_{k,e}\}$, initial model ${\theta}^{0}_{G}$.
\Statex \textbf{Output:} final global model $\boldsymbol{\theta}^{R}_{G}$.
\Statex \textbf{Server Executes:}
\State Initialize $\bar{\mathbf{F}}^{0}_{G} \gets \mathbf{0}$
\For{each experience $e$:}
\For{$r = 1$ to $R$}
    \State Select clients $\mathcal{S}_{r}$
    \State Send $\boldsymbol{\theta}^{r-1}_{G}$ to clients in $\mathcal{S}_{r}$

    \For{each client $k \in \mathcal{S}_{r}$ \textbf{in parallel}}
        \If{$\bar{\mathbf{F}}_{G}$ was updated in ${r}-{1}$ round}
                \State ${\theta}^{r}_{K},\bar{{F}}^r_{K} \gets \textbf{ClientUpdate}           ({\theta}^{r-1}_{G},\bar{{F}}_{G})$
        \Else
            \State ${\theta}^{r}_{K}, \bar{{F}}^{r}_{K} \gets \textbf{ClientUpdate}           ({\theta}^{r-1}_{G})$
    \EndIf
    \EndFor
    \State $\boldsymbol{\theta}^{r}_{G} \gets
\sum_{k \in S_r}
\frac{|\mathcal{D}_{k,e}|}
{\sum_{j \in S_r} |\mathcal{D}_{j,e}|}
\boldsymbol{\theta}^{r}_{k}$ 
    \If{$r \bmod I = 0$}
        \State update $\bar{{F}}_{G}$ using  Eq.~\eqref{eq:ours_global_fisher}
    \EndIf
\EndFor
\EndFor
\end{algorithmic}
\begin{algorithmic}[1]
\vspace{0.2cm}
\Statex \textbf{ClientUpdate:}($\boldsymbol{\theta}^{R}_{G}$,$\bar{{F}}_{G}$)
    \If{$\bar{F}_{G}$ was received}
        \State $\bar{F}^r_{K} \gets \bar{F}_{G}$ 
    \EndIf
    \State optimize Eq.~\eqref{eq:ewc_loss} on local Data $D_{k,e}$ to get updated local model $\theta^{r}_{k}$
    \If {${r} \bmod {I}={0}$}
    \State calculate new ${F}^r_{k}$ using Eq.~\eqref{eq:dr_fisher} and transmit to server.
    \EndIf 
    
\end{algorithmic}
\end{algorithm}

\begin{figure}[htbp]
\centerline{\includegraphics[width=\columnwidth]{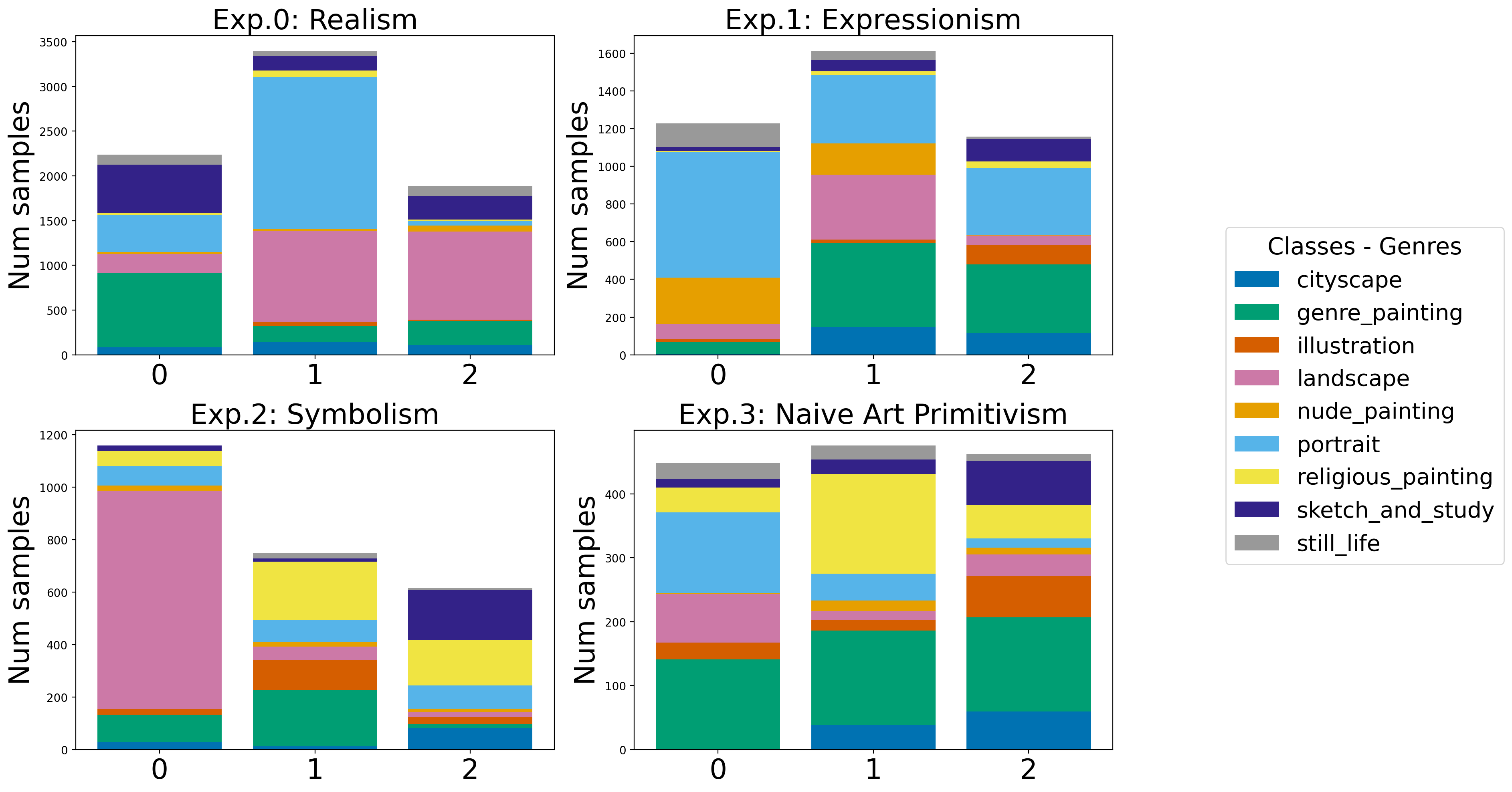}}
\caption{Visualisation of the dataset partitioning with Dirichlet partitioner ($\alpha=1$) among three clients for each experience (style).}
\label{fig:partitioning iid}
\end{figure}

\section{Evaluation}
\subsection{Dataset Preparation}
Artistic styles evolve over time, introducing natural domain shifts in real-world visual data. This requires AI models to adapt to emerging styles while retaining previously learned knowledge, making it a practical continual learning problem. Motivated by this scenario, we evaluate our method on WikiArt, a large-scale digital art dataset spanning diverse artists, genres, and artistic styles. We use the refined version introduced by Tan et al.~\cite{artgan2018}, accessed through the Hugging Face Datasets Hub. The dataset contains paintings from 11 genres, such as \textit{landscape} and \textit{portrait}, and 27 styles, such as \textit{Impressionism} and \textit{Realism}.

We define a genre classification task and construct a domain-incremental scenario by sequentially introducing artistic styles as experiences, while keeping the genre classification task fixed. WikiArt has a sparse style distribution across genres, with many styles missing certain genres. To create a clear domain-incremental scenario and ensure adequate samples per client and per experience, we drop two genres: \textit{Unknown Genre}, due to vague semantics, and \textit{Abstract Painting}, as it is mainly represented by a single style. We then select four styles that cover all nine remaining genres, forming four experiences as follows: \textit{Exp.~0: Realism}, \textit{Exp.~1: Expressionism}, \textit{Exp.~2: Symbolism}, and \textit{Exp.~3: Naive Art Primitivism}. The resulting subset contains approximately 20k images in total.

We partition each experience style among three clients using a Dirichlet distribution to create non-IID client partitions. The resulting distributions are shown in Fig.~\ref{fig:partitioning iid}. We keep 20\% of each partition as a local validation dataset.

\subsection{Experimental setup}
We compare FedCurv-DR with the FedAvg baseline and the original FedCurv. We implement FedCurv with both EWC and EWC-DR importance estimates, and evaluate FedCurv-DR using two different intervals $I$. 
We use a pretrained EfficientNet-B0 model from torchvision as the backbone, with approximately 5.3 million parameters pretrained on ImageNet. For training, we use Adam optimizer with a learning rate of 0.0005, a batch size of 16, and no weight decay. Clients train for 2 local epochs per round, with 5 rounds per experience and 20 rounds in total. We set regularisation strength to $\lambda=200$. We use the same training settings for all methods and experiences to ensure a fair comparison. Our goal is not to achieve optimal performance, but rather to evaluate knowledge retention and convergence stability.
All experiments are run in simulation on a single machine equipped with an 8GB NVIDIA RTX 4070 GPU. We use the Flower AI framework~\cite{beutel_flower_2022}, combined with continual learning strategies from Avalanche~\cite{lomonaco_avalanche_2021} and the CodeCarbon framework~\cite{benoit_courty_2024_11171501} for energy tracking.

\begin{table*}[!t]
\caption{Summary of Final Accuracy, Backward Transfer, Disparity, and Energy Consumption Across Methods}
\label{tab:run_summary_metrics}
\begin{center}
\scriptsize
\begin{tabular}{|c|c|c|c|c|}
\hline
\textbf{Method} &
\textbf{Final Stream Acc.(\%) $\uparrow$} &
\textbf{BWT(pp) $\uparrow$} &
\textbf{Disparity(pp) $\downarrow$} &
\textbf{Energy(kWh) $\downarrow$} \\
\hline
FedAvg &
63.18 &
-11.30 &
7.22 &
\textbf{0.0296} \\
\hline
FedCurv(EWC) &
63.99 &
-12.32 &
3.09 &
0.0448 \\
\hline
FedCurv(EWC-DR) &
65.46 &
-10.26 &
4.56 &
0.0458 \\
\hline
FedCurv-DR(interval=2, beta=0.95) &
\textbf{66.92} &
\textbf{-6.92} &
\textbf{2.89} &
0.0385 \\
\hline
FedCurv-DR(interval=5, beta=0.95) &
65.86 &
-7.49 &
5.91 &
0.0334 \\
\hline
\end{tabular}
\end{center}
\end{table*}

\begin{figure*}[!t]
\centering

\begin{minipage}{0.32\textwidth}
    \centering
    \includegraphics[width=\linewidth]{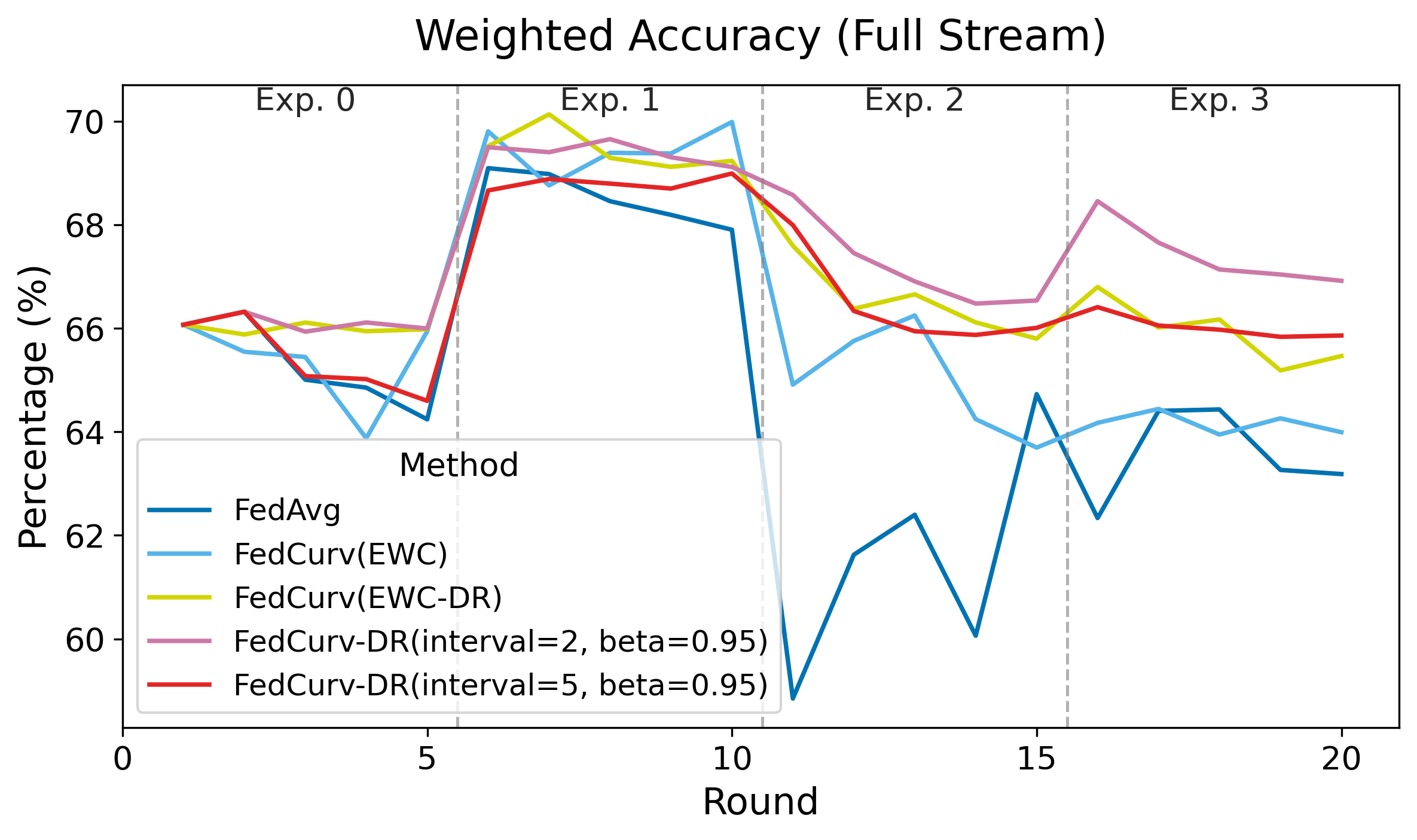}
    \vspace{-2mm}
    
    \textbf{(a)} 
\end{minipage}
\hfill
\begin{minipage}{0.32\textwidth}
    \centering
    \includegraphics[width=\linewidth]{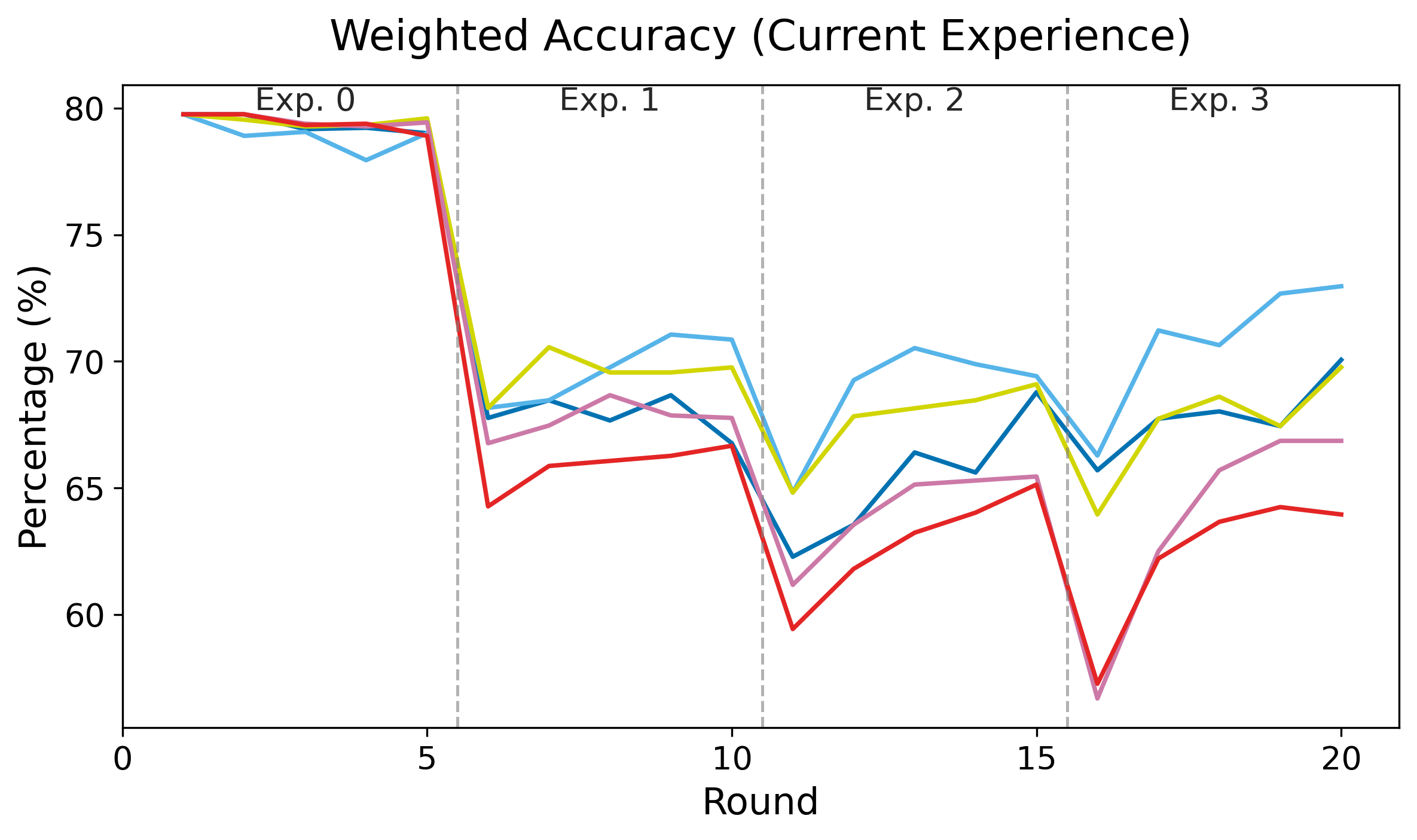}
    \vspace{-2mm}
    
    \textbf{(b)} 
\end{minipage}
\hfill
\begin{minipage}{0.32\textwidth}
    \centering
    \includegraphics[width=\linewidth]{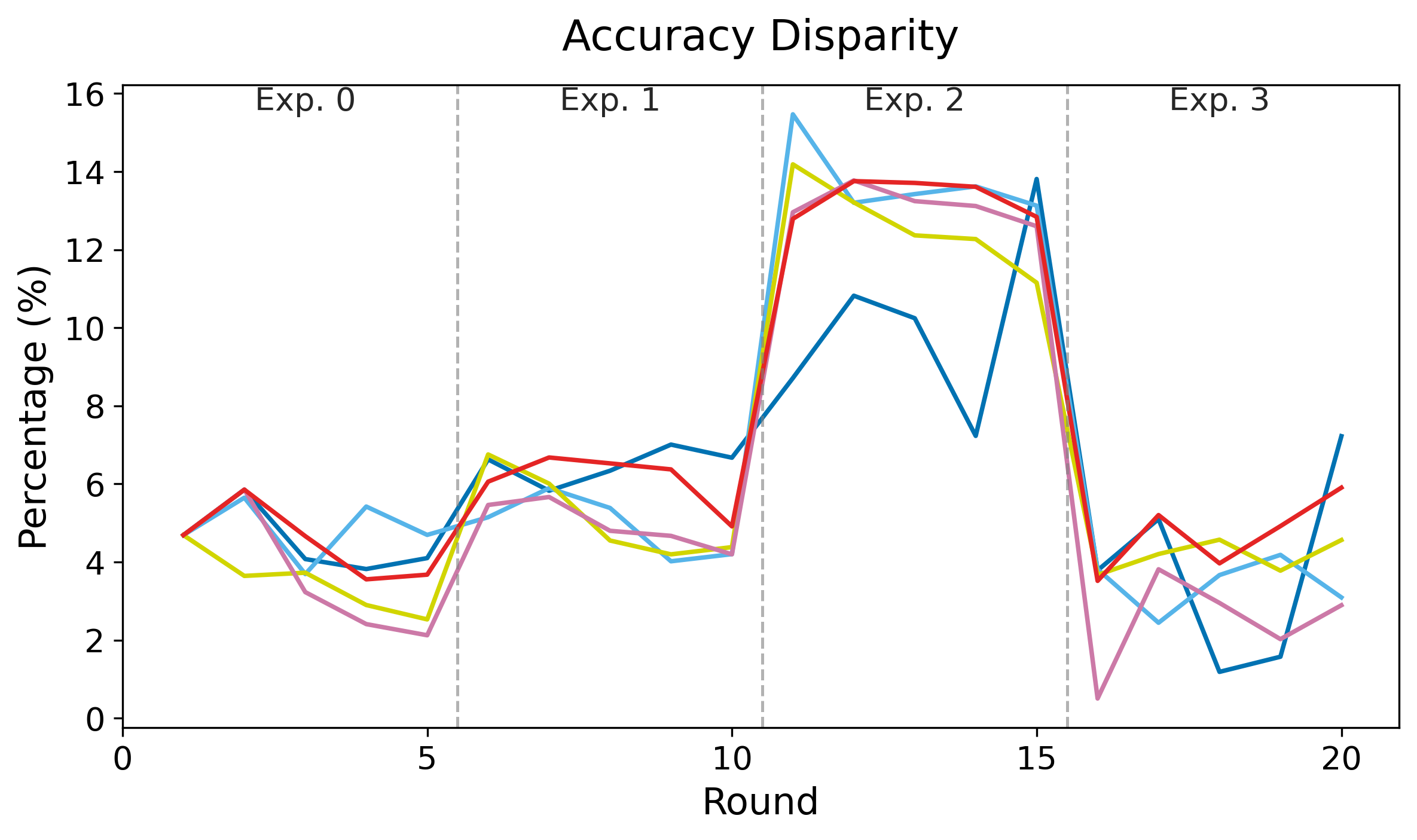}
    \vspace{-2mm}
    
    \textbf{(c)}
\end{minipage}

\caption{Comparison of federated continual learning strategies across communication rounds. 
(a) Full-stream accuracy evaluates retention over all learned experiences. 
(b) Current-experience accuracy measures performance on the active experience. 
(c) Accuracy disparity reports client-level performance differences as a proxy for fairness.}
\label{fig:main_results}
\end{figure*}

\subsection{Evaluation metrics}
For performance evaluation, we use weighted average accuracy and report it both per experience and over the full stream. To quantify catastrophic forgetting mitigation, we use the Backward Transfer (BWT)~\cite{lopez-paz_gradient_2022} metric , defined as
\begin{equation}
\mathrm{BWT}
=
\frac{1}{T-1}
\sum_{i=1}^{T-1}
\left(
A_{T,i} - A_{i,i}
\right),
\label{eq:bwt}
\end{equation}
where $A_{j,i}$ denotes the accuracy on experience $i$ after learning experience $j$, and $T$ is the total number of experiences.

To quantify performance disparity across clients, we compute the standard deviation of client accuracies:
\begin{equation}
\mathrm{Disparity}
=
\sqrt{
\frac{1}{|\mathcal{C}|-1}
\sum_{k \in \mathcal{C}}
\left(
A_k - \bar{A}
\right)^2
},
\label{eq:accuracy_disparity}
\end{equation}
where $\mathcal{C}$ is the set of clients, $A_k$ is the accuracy of client $k$, and $\bar{A}$ is the average client accuracy. This follows the client-level disparity formulation suggested by~\cite{zhang_unified_2024}. Finally, we report the energy consumption and equivalent carbon emissions for each experiment using the CodeCarbon framework.

\section{Results}

\subsection{Performance}

Fig.~\ref{fig:main_results} and Table~\ref{tab:run_summary_metrics} show that FedCurv-DR mitigates catastrophic forgetting more effectively than FedAvg. FedAvg exhibits clear drops in full-stream accuracy after experience transitions and obtains a BWT of -11.30 pp. In contrast, FedCurv-DR with intervals of 2 and 5 rounds achieves higher final stream accuracy, 66.92\% and 65.86\%, compared with 63.18\% for FedAvg, while also improving BWT to -6.92 pp and -7.49 pp, respectively. This indicates that the higher stream accuracy of FedCurv-DR is mainly due to better retention of previous experiences. Current-experience accuracy remains competitive, suggesting that the regularisation improves stability without severely limiting adaptation to new data.

\subsection{Fairness}

We use client-level disparity as a proxy for fairness. FedAvg reaches a final disparity of 7.22 pp, while all FedCurv-based methods reduce this value. The lowest disparity is achieved by FedCurv-DR with interval 2, at 2.89 pp, followed by FedCurv(EWC) at 3.09 pp. Current results are not conclusive on fairness but they suggest that regularisation methods can support more balanced performance across heterogeneous clients.

\subsection{Energy Efficiency}

Table~\ref{tab:run_summary_metrics} reports energy consumption. FedAvg has the lowest energy use, 0.0296 kWh, but also the weakest retention. FedCurv(EWC) and FedCurv(EWC-DR) require higher energy due to importances calculation, 0.0448 kWh and 0.0458 kWh. In contrast, the interval-based variants reduce this overhead: interval 2 consumes 0.0385 kWh, while interval 5 further reduces energy to 0.0334 kWh with only a small drop in final stream accuracy.

\section{Conclusion and Future work}

In this paper, we proposed a lightweight approach for inclusive and lightweight Federated Continual Learning. The results show that the proposed method mitigates catastrophic forgetting while reducing the energy footprint through interval-based importance updates. We also observed improved client-level fairness in heterogeneous settings, an important property for bias mitigation in cultural heritage scenarios where institutions may differ in collection size, data quality, and style distribution. A limitation of this study is that the experiments use a public dataset synthetically partitioned across three simulated clients. This controlled setup allow us to isolate the effects of forgetting, client disparity, and energy consumption, but does not fully capture real institutional deployment constraints. FedCurv-DR is directly compatible with deployment-level privacy mechanisms; practical systems should combine it with secure aggregation~\cite{bonawitz_practical_2016} and differential privacy~\cite{abadi_deep_2016} to bound information leakage from model updates. In real institutional deployments, the FL server would require operation by a trusted neutral party, such as a national heritage agency or research infrastructure, with defined governance procedures covering leakage auditing, intellectual property attribution, and licensing of contributed collections, in alignment with European AI and cultural heritage data policies  \cite{doi/10.2804/4225375}. 
Future work will evaluate larger-scale and real distributed settings, together with ablation studies on $I$, $\beta$, and $\lambda$.

\section*{Acknowledgment}

This work is funded by the European Union's Horizon Europe Research and
Innovation Programme through the RAIDO project (Grant Agreement No. 101135800).

\bibliographystyle{IEEEtran}
\bibliography{continual}

@inproceedings{abadi_deep_2016,
	title = {Deep Learning with Differential Privacy},
	url = {http://arxiv.org/abs/1607.00133},
	doi = {10.1145/2976749.2978318},
	pages = {308--318},
	booktitle = {Proceedings of the 2016 {ACM} {SIGSAC} Conference on Computer and Communications Security},
	author = {Abadi, Martín and others},
	urldate = {2026-06-05},
	date = {2016-10-24},
	eprinttype = {arxiv},
	eprint = {1607.00133 [stat.ML]},
}

@book{doi/10.2804/4225375,
author = {{European Data Protection Supervisor}},
title = {AI Act Regulation (EU) 2024/1689 – Regulation (EU) 2024/1689 of the European Parliament and of the Council of 13 June 2024 laying down harmonised rules on artificial intelligence and amending Regulations (EC) No 300/2008, (EU) No 167/2013, (EU) No 168/2013, (EU) 2018/858, (EU) 2018/1139 and (EU) 2019/2144 and Directives 2014/90/EU, (EU) 2016/797 and (EU) 2020/1828 (Artificial Intelligence Act) (Text with EEA relevance)},
publisher = {Publications Office of the European Union},
year = {2025},
doi = {doi/10.2804/4225375}}

@inproceedings{adorni_towards_2025,
	title = {Towards a Manifesto for Cyber Humanities: Paradigms, Ethics, and Prospects},
	url = {https://ieeexplore.ieee.org/abstract/document/11279659},
	doi = {10.1109/IEEE-CH65308.2025.11279659},
	shorttitle = {Towards a Manifesto for Cyber Humanities},
	eventtitle = {2025 {IEEE} International Conference on Cyber Humanities ({IEEE}-{CH})},
	pages = {1--8},
	booktitle = {2025 {IEEE} International Conference on Cyber Humanities ({IEEE}-{CH})},
	author = {Adorni, Giovanni and Bellini, Emanuele},
	urldate = {2026-06-04},
	date = {2025-09},
}

@article{bellini_cyber_2025,
	title = {Cyber Humanities for Heritage Security},
	volume = {68},
	issn = {0001-0782},
	url = {https://dl.acm.org/doi/10.1145/3735659},
	doi = {10.1145/3735659},
	pages = {112--117},
	number = {12},
	journaltitle = {Commun. {ACM}},
	author = {Bellini, Emanuele},
	urldate = {2026-06-04},
	date = {2025-08-25},
}

@software{benoit_courty_2024_11171501,
  author       = {Courty,Benoit and others},
  title        = {mlco2/codecarbon: v2.4.1},
  month        = may,
  year         = 2024,
  publisher    = {Zenodo},
  version      = {v2.4.1},
  doi          = {10.5281/zenodo.11171501},
  url          = {https://doi.org/10.5281/zenodo.11171501}
}

@misc{lopez-paz_gradient_2022,
	title = {Gradient Episodic Memory for Continual Learning},
	url = {http://arxiv.org/abs/1706.08840},
	doi = {10.48550/arXiv.1706.08840},
	number = {{arXiv}:1706.08840},
	publisher = {{arXiv}},
	author = {Lopez-Paz, David and Ranzato, Marc'Aurelio},
	urldate = {2026-05-09},
	date = {2022-09-13},
	eprinttype = {arxiv},
	eprint = {1706.08840 [cs]},
}

@inproceedings{rolnick_experience_2019,
  title     = {Experience Replay for Continual Learning},
  author    = {Rolnick, David and Ahuja, Arun and Schwarz, Jonathan and Lillicrap, Timothy and Wayne, Gregory},
  booktitle = {Advances in Neural Information Processing Systems},
  volume    = {32},
  year      = {2019},
  publisher = {Curran Associates, Inc.},
}

@techreport{eprs2023aiheritage,
  author = {{European Parliamentary Research Service}},
  title = {Artificial intelligence in the context of cultural heritage and museums},
  institution = {European Parliament},
  year = {2023}
}

@article{disli2025copyright,
  author = {Dişli, Melike},
  title = {Copyright and Licencing for Cultural Heritage Collections As Data},
  journal = {Journal of Open Humanities Data},
  volume = {11},
  year = {2025},
  doi = {10.5334/johd.263}
}

@article{artgan2018,
title={Improved ArtGAN for Conditional Synthesis of Natural Image and Artwork},
author={Tan, Wei Ren and Chan, Chee Seng and Aguirre, Hernan and Tanaka, Kiyoshi},
journal={IEEE Transactions on Image Processing},
volume = {28},
number = {1},
pages = {394--409},
year = {2019},
url = {https://doi.org/10.1109/TIP.2018.2866698},
doi = {10.1109/TIP.2018.2866698}
}

@article{foka_tracing_2025,
	title = {Tracing the bias loop: {AI}, cultural heritage and bias-mitigating in practice},
	volume = {40},
	issn = {1435-5655},
	url = {https://doi.org/10.1007/s00146-025-02349-z},
	doi = {10.1007/s00146-025-02349-z},
	shorttitle = {Tracing the bias loop},
	pages = {5835--5847},
	number = {8},
	journaltitle = {{AI} \& Soc},
	author = {Foka, Anna and Griffin, Gabriele and Ortiz Pablo, Dalia and Rajkowska, Paulina and Badri, Sushruth},
	urldate = {2026-04-22},
	date = {2025-12-01},
	langid = {english},
}

@unpublished{tziolas_energy-efficient_2026,
	location = {Florence, Italy},
	title = {Energy-Efficient and Privacy-Preserving Federated Continual Learning for Cultural Heritage Preservation and Digital Humanities},
	url = {https://zenodo.org/records/18504685},
	doi = {10.1109/IEEE-CH65308.2025.11279476},
	author = {Tziolas, George and others},
	urldate = {2026-05-09},
	date = {2026-01-20},
	note = {Conference Name: 2025 {IEEE} International Conference on Cyber Humanities ({IEEE}-{CH}) ({IEEE}-{CH}2025)
{ISBN}: 9798331514365
Publisher: {IEEE}},
}

@article{fiorucci_machine_2020,
	title = {Machine Learning for Cultural Heritage: A Survey},
	volume = {133},
	issn = {0167-8655},
	url = {https://www.sciencedirect.com/science/article/pii/S0167865520300532},
	doi = {10.1016/j.patrec.2020.02.017},
	shorttitle = {Machine Learning for Cultural Heritage},
	pages = {102--108},
	journaltitle = {Pattern Recognition Letters},
	author = {Fiorucci, Marco and Khoroshiltseva, Marina and Pontil, Massimiliano and Traviglia, Arianna and Del Bue, Alessio and James, Stuart},
	urldate = {2026-05-09},
	date = {2020-05-01},
}

@article{zhitomirsky-geffet_what_2022,
	title = {What do they make us see: a comparative study of cultural bias in online databases of two large museums},
	volume = {79},
	doi = {10.1108/JD-02-2022-0047},
	shorttitle = {What do they make us see},
	journaltitle = {Journal of Documentation},
	author = {Zhitomirsky-Geffet, Maayan and Kizhner, Inna and Minster, Sara},
	date = {2022-07-01},
}

@article{schwartz2020greenai,
  title   = {Green AI},
  author  = {Schwartz, Roy and Dodge, Jesse and Smith, Noah A. and Etzioni, Oren},
  journal = {Communications of the ACM},
  volume  = {63},
  number  = {12},
  pages   = {54--63},
  year    = {2020},
  doi     = {10.1145/3381831}
}

@misc{ezzeldin_fairfed_2022,
	title = {{FairFed}: {Enabling} {Group} {Fairness} in {Federated} {Learning}},
	shorttitle = {{FairFed}},
	url = {http://arxiv.org/abs/2110.00857},
	doi = {10.48550/arXiv.2110.00857},
	urldate = {2026-04-20},
	publisher = {arXiv},
	author = {Ezzeldin, Yahya H. and Yan, Shen and He, Chaoyang and Ferrara, Emilio and Avestimehr, Salman},
	month = nov,
	year = {2022},
	note = {arXiv:2110.00857 [cs]},
}

@inproceedings{lo_fairwag_2025,
	address = {New York, NY, USA},
	series = {{EAAMO} '25},
	title = {{FairWAG}: {Fairness}-aware {Weighted} {Aggregation} for {Graph} {Learning} in a {Federated} {Setting}},
	isbn = {979-8-4007-2140-3},
	shorttitle = {{FairWAG}},
	url = {https://dl.acm.org/doi/10.1145/3757887.3763013},
	doi = {10.1145/3757887.3763013},
	urldate = {2026-04-20},
	booktitle = {Proceedings of the 5th {ACM} {Conference} on {Equity} and {Access} in {Algorithms}, {Mechanisms}, and {Optimization}},
	publisher = {Association for Computing Machinery},
	author = {Lo, Kuan-Chieh and He, Yuntian and Jiang, Yuze and Parthasarathy, Srinivasan},
	month = aug,
	year = {2025},
	pages = {119--150},
}

@misc{zhang_target_2023,
	title = {{TARGET}: {Federated} {Class}-{Continual} {Learning} via {Exemplar}-{Free} {Distillation}},
	shorttitle = {{TARGET}},
	url = {http://arxiv.org/abs/2303.06937},
	doi = {10.48550/arXiv.2303.06937},
	urldate = {2025-06-16},
	publisher = {arXiv},
	author = {Zhang, Jie and Chen, Chen and Zhuang, Weiming and Lv, Lingjuan},
	month = aug,
	year = {2023},
	note = {arXiv:2303.06937 [cs]},
}

@inproceedings{yao_continual_2020,
	address = {Abu Dhabi, United Arab Emirates},
	title = {Continual {Local} {Training} {For} {Better} {Initialization} {Of} {Federated} {Models}},
	copyright = {https://ieeexplore.ieee.org/Xplorehelp/downloads/license-information/IEEE.html},
	isbn = {978-1-7281-6395-6},
	url = {https://ieeexplore.ieee.org/document/9190968/},
	doi = {10.1109/ICIP40778.2020.9190968},
	language = {en},
	urldate = {2025-06-16},
	booktitle = {2020 {IEEE} {International} {Conference} on {Image} {Processing} ({ICIP})},
	publisher = {IEEE},
	author = {Yao, Xin and Sun, Lifeng},
	month = oct,
	year = {2020},
	pages = {1736--1740},
}

@misc{usmanova_federated_2022,
	title = {Federated {Continual} {Learning} through distillation in pervasive computing},
	url = {http://arxiv.org/abs/2207.08181},
	doi = {10.48550/arXiv.2207.08181},
	urldate = {2025-04-25},
	publisher = {arXiv},
	author = {Usmanova, Anastasiia and Portet, François and Lalanda, Philippe and Vega, German},
	month = jul,
	year = {2022},
	note = {arXiv:2207.08181 [cs]},
}

@article{asad_fedopt_2020,
	title = {{FedOpt}: {Towards} {Communication} {Efficiency} and {Privacy} {Preservation} in {Federated} {Learning}},
	volume = {10},
	copyright = {http://creativecommons.org/licenses/by/3.0/},
	issn = {2076-3417},
	shorttitle = {{FedOpt}},
	url = {https://www.mdpi.com/2076-3417/10/8/2864},
	doi = {10.3390/app10082864},
	language = {en},
	number = {8},
	urldate = {2025-04-14},
	journal = {Applied Sciences},
	publisher = {Multidisciplinary Digital Publishing Institute},
	author = {Asad, Muhammad and Moustafa, Ahmed and Ito, Takayuki},
	month = jan,
	year = {2020},
	note = {Number: 8},
	pages = {2864},
}

@misc{li_fair_2020,
	title = {Fair {Resource} {Allocation} in {Federated} {Learning}},
	url = {http://arxiv.org/abs/1905.10497},
	doi = {10.48550/arXiv.1905.10497},
	urldate = {2025-04-14},
	publisher = {arXiv},
	author = {Li, Tian and Sanjabi, Maziar and Beirami, Ahmad and Smith, Virginia},
	month = feb,
	year = {2020},
	note = {arXiv:1905.10497 [cs]},
}

@unpublished{beutel_flower_2022,
	title = {{FLOWER}: {A} {FRIENDLY} {FEDERATED} {LEARNING} {FRAMEWORK}},
	shorttitle = {{FLOWER}},
	url = {https://hal.science/hal-03601230},
	urldate = {2025-03-10},
	author = {Beutel, Daniel J. and others},
	month = nov,
	year = {2022},
}

@misc{bonawitz_practical_2016,
	title = {Practical {Secure} {Aggregation} for {Federated} {Learning} on {User}-{Held} {Data}},
	url = {http://arxiv.org/abs/1611.04482},
	doi = {10.48550/arXiv.1611.04482},
	urldate = {2025-02-10},
	publisher = {arXiv},
	author = {Bonawitz, Keith and others},
	month = nov,
	year = {2016},
	note = {arXiv:1611.04482 [cs]},
}

@article{schoinas_federated_2024,
	title = {Federated {Learning}: {Challenges}, {SoTA}, {Performance} {Improvements} and {Application} {Domains}},
	volume = {5},
	copyright = {https://creativecommons.org/licenses/by/4.0/legalcode},
	issn = {2644-125X},
	shorttitle = {Federated {Learning}},
	url = {https://ieeexplore.ieee.org/document/10677499/},
	doi = {10.1109/OJCOMS.2024.3458088},
	language = {en},
	urldate = {2025-02-05},
	journal = {IEEE Open Journal of the Communications Society},
	author = {Schoinas, Ioannis and others},
	year = {2024},
	pages = {5933--6017},
}

@misc{mcmahan_communication-efficient_2023,
	title = {Communication-{Efficient} {Learning} of {Deep} {Networks} from {Decentralized} {Data}},
	url = {http://arxiv.org/abs/1602.05629},
	doi = {10.48550/arXiv.1602.05629},
	urldate = {2025-02-05},
	publisher = {arXiv},
	author = {McMahan, H. Brendan and Moore, Eider and Ramage, Daniel and Hampson, Seth and Arcas, Blaise Agüera y},
	month = feb,
	year = {2017},
	note = {arXiv:1602.05629 [cs]},
}

@article{kirkpatrick_overcoming_2017,
	title = {Overcoming catastrophic forgetting in neural networks},
	volume = {114},
	issn = {0027-8424, 1091-6490},
	url = {http://arxiv.org/abs/1612.00796},
	doi = {10.1073/pnas.1611835114},
	number = {13},
	urldate = {2025-06-20},
	journal = {Proceedings of the National Academy of Sciences},
	author = {Kirkpatrick, James and others},
	month = mar,
	year = {2017},
	note = {arXiv:1612.00796 [cs]},
	pages = {3521--3526},
}

@misc{shoham_overcoming_2019,
	title = {Overcoming {Forgetting} in {Federated} {Learning} on {Non}-{IID} {Data}},
	url = {http://arxiv.org/abs/1910.07796},
	doi = {10.48550/arXiv.1910.07796},
	urldate = {2025-07-09},
	publisher = {arXiv},
	author = {Shoham, Neta and Avidor, Tomer and Keren, Aviv and Israel, Nadav and Benditkis, Daniel and Mor-Yosef, Liron and Zeitak, Itai},
	month = oct,
	year = {2019},
	note = {arXiv:1910.07796 [cs]},
}

@misc{yoon_federated_2021,
	title = {Federated {Continual} {Learning} with {Weighted} {Inter}-client {Transfer}},
	url = {http://arxiv.org/abs/2003.03196},
	doi = {10.48550/arXiv.2003.03196},
	urldate = {2025-07-11},
	publisher = {arXiv},
	author = {Yoon, Jaehong and Jeong, Wonyong and Lee, Giwoong and Yang, Eunho and Hwang, Sung Ju},
	month = jun,
	year = {2021},
	note = {arXiv:2003.03196 [cs]},
}

@article{zhang_towards_2025,
	title = {Towards fairness-aware and privacy-preserving enhanced collaborative learning for healthcare},
	volume = {16},
	copyright = {2025 The Author(s)},
	issn = {2041-1723},
	url = {https://www.nature.com/articles/s41467-025-58055-3},
	doi = {10.1038/s41467-025-58055-3},
	language = {en},
	number = {1},
	urldate = {2025-07-22},
	journal = {Nature Communications},
	publisher = {Nature Publishing Group},
	author = {Zhang, Feilong and Zhai, Deming and Bai, Guo and Jiang, Junjun and Ye, Qixiang and Ji, Xiangyang and Liu, Xianming},
	month = mar,
	year = {2025},
	pages = {2852},
}

@article{zhang_unified_2024,
	title = {Unified fair federated learning for digital healthcare},
	volume = {5},
	issn = {2666-3899},
	url = {https://www.sciencedirect.com/science/article/pii/S2666389923003148},
	doi = {10.1016/j.patter.2023.100907},
	number = {1},
	urldate = {2025-09-02},
	journal = {Patterns},
	author = {Zhang, Fengda and Shuai, Zitao and Kuang, Kun and Wu, Fei and Zhuang, Yueting and Xiao, Jun},
	month = jan,
	year = {2024},
	pages = {100907},
}

@misc{lin_ensemble_2021,
	title = {Ensemble {Distillation} for {Robust} {Model} {Fusion} in {Federated} {Learning}},
	url = {http://arxiv.org/abs/2006.07242},
	doi = {10.48550/arXiv.2006.07242},
	urldate = {2025-09-03},
	publisher = {arXiv},
	author = {Lin, Tao and Kong, Lingjing and Stich, Sebastian U. and Jaggi, Martin},
	month = mar,
	year = {2021},
	note = {arXiv:2006.07242 [cs]},
}

@misc{li_federated_2020,
	title = {Federated {Optimization} in {Heterogeneous} {Networks}},
	url = {http://arxiv.org/abs/1812.06127},
	doi = {10.48550/arXiv.1812.06127},
	urldate = {2025-10-29},
	publisher = {arXiv},
	author = {Li, Tian and Sahu, Anit Kumar and Zaheer, Manzil and Sanjabi, Maziar and Talwalkar, Ameet and Smith, Virginia},
	month = apr,
	year = {2020},
	note = {arXiv:1812.06127 [cs]},
}

@misc{ven_three_2019,
	title = {Three scenarios for continual learning},
	url = {http://arxiv.org/abs/1904.07734},
	doi = {10.48550/arXiv.1904.07734},
	language = {en},
	urldate = {2025-12-18},
	publisher = {arXiv},
	author = {Ven, Gido M. van de and Tolias, Andreas S.},
	month = apr,
	year = {2019},
	note = {arXiv:1904.07734 [cs]},
}

@misc{lomonaco_avalanche_2021,
	title = {Avalanche: an {End}-to-{End} {Library} for {Continual} {Learning}},
	shorttitle = {Avalanche},
	url = {http://arxiv.org/abs/2104.00405},
	doi = {10.48550/arXiv.2104.00405},
	urldate = {2026-03-10},
	publisher = {arXiv},
	author = {Lomonaco, Vincenzo and others},
	month = apr,
	year = {2021},
	note = {arXiv:2104.00405 [cs]},
}

@misc{aljundi_memory_2018,
	title = {Memory {Aware} {Synapses}: {Learning} what (not) to forget},
	shorttitle = {Memory {Aware} {Synapses}},
	url = {http://arxiv.org/abs/1711.09601},
	doi = {10.48550/arXiv.1711.09601},
	urldate = {2026-04-15},
	publisher = {arXiv},
	author = {Aljundi, Rahaf and Babiloni, Francesca and Elhoseiny, Mohamed and Rohrbach, Marcus and Tuytelaars, Tinne},
	month = oct,
	year = {2018},
	note = {arXiv:1711.09601 [cs]},
}

@misc{zenke_continual_2017,
	title = {Continual {Learning} {Through} {Synaptic} {Intelligence}},
	url = {http://arxiv.org/abs/1703.04200},
	doi = {10.48550/arXiv.1703.04200},
	urldate = {2026-04-15},
	publisher = {arXiv},
	author = {Zenke, Friedemann and Poole, Ben and Ganguli, Surya},
	month = jun,
	year = {2017},
	note = {arXiv:1703.04200 [cs]},
}

@misc{liu_elastic_2026,
	title = {Elastic Weight Consolidation Done Right for Continual Learning},
	rights = {{arXiv}.org perpetual, non-exclusive license},
	url = {https://arxiv.org/abs/2603.18596},
	doi = {10.48550/ARXIV.2603.18596},
	publisher = {{arXiv}},
	author = {Liu, Xuan and Chang, Xiaobin},
	urldate = {2026-05-07},
	date = {2026},
	langid = {english},
	note = {Version Number: 3},
}

\end{document}